%% file: emnlp-ijcnlp-2019.tex
\documentclass[11pt,a4paper]{article}
\usepackage[hyperref]{emnlp-ijcnlp-2019}
\usepackage{times}
\usepackage{amssymb}
\usepackage{mathtools}
\usepackage{bm}
\usepackage{multirow}
\usepackage{graphicx}
\usepackage{subcaption}
\usepackage{float}

\usepackage{url}

\aclfinalcopy % Uncomment this line for the final submission

\title{On the Effectiveness of the Pooling Methods \\ for Biomedical Relation Extraction with Deep Learning}

\author{Tuan Ngo Nguyen$^\dag$, Franck Dernoncourt$^\ddagger$ and Thien Huu Nguyen$^\dag$ \\
    $^\dag$ Department of Computer and Information Science, University of Oregon \\
    $^\ddagger$ Adobe Research \\
    {\tt \{tnguyen, thien\}@cs.uoregon.edu, dernonco@adobe.com} }

\date{}

\begin{document}
\maketitle
\begin{abstract}
  Deep learning models have achieved state-of-the-art performances on many relation extraction datasets. A common element in these deep learning models involves the pooling mechanisms where a sequence of hidden vectors is aggregated to generate a single representation vector, serving as the features to perform prediction for RE. Unfortunately, the models in the literature tend to employ different strategies to perform pooling for RE, leading to the challenge to determine the best pooling mechanism for this problem, especially in the biomedical domain. In order to answer this question, in this work, we conduct a comprehensive study to evaluate the effectiveness of different pooling mechanisms for the deep learning models in biomedical RE. The experimental results suggest that dependency-based pooling is the best pooling strategy for RE in the biomedical domain, yielding the state-of-the-art performance on two benchmark datasets for this problem.
\end{abstract}

\input{introduction.tex}

\input{models.tex}

\input{experiments.tex}

\input{related_conclusion.tex}

\bibliography{emnlp-ijcnlp-2019}
\bibliographystyle{acl_natbib}

\end{document}

%% file: introduction.tex
\section{Introduction}

% introduce relation extraction and biomedical domain
%Relation extraction (RE) is a natural language processing task of identifying the semantic relation holding between two entity mentions in text. There is tremendous value in automating extraction of key discoveries in biomedical scientific publications, e.g., drug-drug interaction, drug-disease relation. Accordingly, number of RE research in biomedical domain has been growing every year --either feature-based kernels \cite{raihani2017Rich, chowdhury2013FBKirst, lever2016VERSE} or deep learning \cite{mehryary2016Deep, bjorne2018Biomedical, nguyen2018Convolutional}

In order to analyze the entities in text, it is crucial to understand how the entities are related to each other in the documents. In the literature, this problem is formalized as relation extraction (RE), an important task in information extraction. RE aims to identify the semantic relationships between two entity mentions within the same sentences in text. Due to its important applications on many areas of natural language processing (e.g., question answering, knowledge base construction), RE has been actively studied in the last decade, featuring a variety of feature-based or kernel-based models for this problem \cite{zelenko2002Kernel,zhou2005Exploring,bunescu2005shortest,Sun:11,Chan:10,Nguyen:09}. Recently, the introduction of deep learning has produced a new generation of models for RE with the state-of-the-art performance on many different benchmark datasets \cite{zeng2014Relation,dossantos2015Classifying,xu2015Classifying,liu2015dependency,zhou2016AttentionBased,wang2016Relation,zhang2017Positionaware,zhang2018Graph}. The advantage of deep learning over the previous approaches for RE is the ability to automatically learn effective features for the sentences from data via various network architectures. The same trend has also been observed for RE in the biomedical domain where deep learning is gaining more and more attention from the research community \cite{mehryary2016Deep,bjorne2018Biomedical, nguyen2018Convolutional,verga2018Simultaneously}.

The typical deep learning models for RE have involved Convolutional Neural Networks (CNN) \cite{zeng2014Relation,nguyen2015Relation,zeng2015Distant,lin2016Neural,zeng2017Incorporating}, Recurrent Neural Networks (RNN), \cite{miwa2016EndtoEnd,zhang2017Positionaware}, Transformer (self-attention) networks \cite{verga2018Simultaneously}, and Graph Convolutional Neural Networks (GCNN) \cite{zhang2018Graph}. There are two major common components in such deep learning models for RE, i.e., the representation component and the pooling component. First, in the representation component, some deep learning architectures are employed to compute a sequence of vectors to represent an input sentence for RE for which each vector tends to capture the specific context information for a word in that sentence.  Such word-specific representation sequence is then fed into the second pooling component (e.g., max pooling) that aggregates the representation vectors to obtain an overall vector to represent the whole input sentence for the classification problem in RE.

While there have been many work in the literature to compare different deep learning architectures for the representation component, the possible methods for the pooling component of the deep learning models have not been systematically benchmarked for RE in general and for the biomedical domain in particular. Specifically, the prior work on relation extraction with deep learning has only assumed one form of pooling in the model without considering the possible alternatives for this component. In this work, we argue that the pooling mechanisms also have significant impact on the performance of the deep learning models for RE and it is important to understand how well different pooling methods perform in this case. Consequently, in this work, we conduct a comprehensive investigation on the effectiveness of different max pooling methods for the deep learning models of RE, focusing on the biomedical domain as the case study. Our goal is to determine the best pooling methods for the deep learning models in biomedical RE. We also want to emphasize the experiments where the pooling methods are compared in a compatible manner with the same representation components and resources for the biomedical RE models in this work. Such compatible comparison is unfortunately very rare in the current literature about deep learning for RE as new models are being intensively proposed, employing a diversity of options and resources (i.e., pre-trained word embeddings, optimizers, etc.). Therefore, this is actually the first work to compare different pooling methods for deep relation extraction on the same setting.

%Given the increasing complexity of the deep learning models recently proposed for RE with a diversity of options and resources (i.e., pre-trained word embeddings, dropouts, optimizers, etc.), we believe there is a need to properly evaluate the pooling methods to achieve an insights into their effectiveness for biomedical RE. In particular, in this work, we emphasize the experiments where the pooling methods are compared in a compatible manner with the same representation components and resources in the deep learning models for biomedical RE.

In the experiments, we find that syntactic information (i.e., dependency parsing) can be exploited to provide the best pooling strategies for biomedical RE. In fact, our experiments also suggest that it is more beneficial to apply the syntactic information in the pooling component of the deep learning models for biomedical RE than that in the representation component. This is different from most of the prior work on relation extraction that has only employed the syntactic information in the representation component of the deep learning models \cite{xu2016Improved,miwa2016EndtoEnd}. Based on the syntax-based pooling mechanism, we achieve the state-of-the-art performance on two benchmark datasets for biomedical RE.

%% file: models.tex
\section{Model}

Relation Extraction can be seen as a multi-class classification problem that takes a sentence and two entity mentions of interest in that sentence as the input. The goal is to predict the semantic relation between these two entity mentions according to some predefined set of relations. Formally, let $W = [w_1, w_2, \ldots, w_n]$ be the input sentence where $n$ is the number of tokens and $w_i$ is the $i$-th word/token in $W$. As entity mentions can span multiple consecutive words/tokens, let $[s_1, e_1]$ be the span of the first entity mention $M_1$ where $s_1$ and $e_1$ are the indexes for the first and last token of $M_1$ respectively. Similarly, we define $[s_2, e_2]$ as the span for the second entity mention $M_2$. For convenience, we assume that the entity mentions are not nested, i.e., $1 \le s_1 \le e_1 < s_2 \le e_2 \le n$.

\subsection{Input Vector Representation}

In order to encode the positions and the entity types of the two entity mentions in the input sentence, following \cite{zhang2018Graph}, we first replace the tokens in the entity mentions $M_1$ and $M_2$ with the special tokens of format $M_1$-$\textit{Type}_1$ and $M_2$-$\textit{Type}_2$ respectively ($\textit{Type}_1$ and $\textit{Type}_2$ represent the entity types of $M_1$ and $M_2$ respectively). The purpose of this replacement is to help the models to abstract from the specific tokens/words of the entity mentions and only focus on their positions and entity types, the two most important pieces of information of the entity mentions for RE.

Given the enriched input sentence, the first step in the deep learning models for RE is to convert each word in the input sentence into a vector to facilitate the real-valued computation of the models. In this work, the vector $v_i$ for $w_i$ is obtained by concatenating the following two vectors:

1. The word embeddings of $w_i$: The embeddings for the special tokens are initialized randomly while the embeddings for the other words are retrieved from the pre-trained word embedding table provided by the \textit{Word2Vec} toolkit with 300 dimensions \cite{mikolov2013Efficient}. 

2. The embeddings for the part-of-speech (POS) tag of $w_i$ in $W$: We assign a POS tag for each word in the input sentence using the Stanford CoreNLP toolkit. The embedding for each POS tag is also randomly initialized in this case.

Note that both the word embeddings and the POS embeddings are updated during the training time of the models in this work. The word-to-vector conversion transforms the input sentence $W = [w_1, w_2, \ldots, w_n]$ into a sequence of vectors $V = [v_1, v_2, \ldots, v_n]$ (respectively) that would be used as the input for all the deep learning models considered in this work to ensure a compatible comparison. As mentioned in the introduction, the deep learning models for RE involves two major components, i.e., the representation component and the pooling component. We describe the options for such components in the following sections.

\subsection{The Representation Component for RE}
\label{sec:model}

Given the input sequence of vectors $V = [v_1, v_2, \ldots, v_n]$, the next step in the deep learning models for RE is to transform this vector sequence into a more abstract vector sequence $A = [a_1, a_2, \ldots, a_n]$ so $a_i$ would capture the underlying representation for the context information specific to the $i$-th word in the sentence. In this work, we examine the following typical architectures to obtain such an abstract sequence $A$ for $V$:

1. {\it CNN} \cite{zeng2014Relation,nguyen2015Relation,dossantos2015Classifying}: {\it CNN} is one of the early deep learning models for RE. It involves an 1D convolution layer over the input vector sequence $V$ with multiple window sizes for the filters. {\it CNN} produces a sequence of vectors in which each vector capture some $n$-grams specific to a word in the sentence. This sequence of vectors is used as $A$ for our purpose.

2. {\it BiLSTM} \cite{nguyen2015combining}: In {\it BiLSTM}, two Long-short Term Memory Networks (LSTM) are run over the input vector sequence $V$ in the forward and backward direction. The hidden vectors generated at the position $i$ by the two networks are then concatenated to constitute the abstract vector $a_i$ for this position. Due to the recurrent nature, $a_i$ involves the context information over the whole input sentence $W$ although a greater focus is put on the context of the current word.

3. {\it BiLSTM-CNN}: This models resembles the MASS model presented in \cite{le2018Largescale}. It first applies a bidirectional LSTM layer over the input sequence $V$ whose results are further processed by a Convolutional Neural Network (CNN) layer as in {\it CNN}. We also use the output of the CNN layer as the abstract vector sequence $A$ for this model.

%The output of the CNN layer is a sequence of vectors in which each vector capture some $n$-grams specific to a word in the sentence. We use the output of the CNN layer as the abstract vector sequence $A$ in this work.

4. {\it BiLSTM-GCNN} \cite{zhang2018Graph}: Similar to {\it BiLSTM-CNN}, {\it BiLSTM-GCNN} also first employs a bidirectional LSTM network to abstract the input vector sequence $V$. However, in the second step, different from {\it BiLSTM-CNN}, {\it BiLSTM-GCNN} introduces a Graph Convolutional Neural Network (GCNN) layer that consumes the LSTM hidden vectors and augments the representation for a word with the representation vectors of the surrounding words in the dependency trees. The output of the GCNN layer is also a sequence of vectors to represent the contexts for the words in the sentence and functions as the abstract sequence $A$ in our case. {\it BiLSTM-GCNN} \cite{zhang2018Graph} is one of the current state-of-the-art models for RE in the literature.

Note that there are many other variants of such models for RE in the literature \cite{xu2016Improved,zhang2017Positionaware,verga2018Simultaneously}. However, as our goal in this paper is to evaluate different pooling mechanisms for RE, we focus on these standard representation learning methods to avoid the confounding effect of the complicated models, thus better revealing the effectiveness of the pooling methods.

\subsection{The Pooling Component for RE}
\label{sec:pool}

The goal of the pooling component is to aggregate the representation vectors in the abstract sequence $A$ to constitute an overall vector $F$ to represent the whole input sentence $W$ and the two entity mentions of interest (i.e., $F = \operatorname{aggregate}(A)$). The overall representation vector should be able to capture the most important features induced in $A$. The typical method to achieve such aggregation in the RE models is to apply the element-wise max-pooling operation over subsets of vectors in $A$ whose results are combined to obtain the overall representation vector. While there are different methods to select the vector subsets for the max-pooling operation, the prior work for RE has only employed one particular selection method in their deep learning models \cite{nguyen2015combining,zhang2018Graph,le2018Largescale}. This raises the question about the impact of the other subset selection methods for such prior RE models. Can these methods benefit from different pooling mechanisms? What are the best pooling methods for the deep learning models in RE? In order to answer these questions, besides the architectures for the representation component in the previous section, we investigate the following subset selection methods for the pooling component of the RE models in this work:

1. {\it ENT-ONLY}: In this pooling method, we use the subsets of the vectors corresponding to the words in the two entity mentions of interest in $A$ for the max-pooling operations (i.e., $M_1$ with the words in the range $[s_1, e_1]$ and $M_2$ with the words in the range  $[s_2, e_2]$). This is motivated by the utmost importance of the two entity mentions of interest for RE and employed in some prior work \cite{nguyen2015combining,zhang2018Graph}:
  \begin{align*}
    F_{M_1} &= \operatorname{max-pool}{(a_{s_1},a_{s_1+1},\ldots,a_{e_1})} \\
    F_{M_2} &= \operatorname{max-pool}{(a_{s_2},a_{s_2+1},\ldots,a_{e_2})} \\
    F_{ENT-ONLY} &= [F_{M_1}, F_{M_2}]
  \end{align*}

2. {\it ENT-SENT}: Besides the entity mentions, the other context words in the sentence might also involve important information for the relation prediction in RE. For instance, in the sentence ``{\it \underline{Acetazolamide} can elevate \underline{cyclosporine} levels.}'', the context word ``{\it elevate}'' is crucial to determine the semantic relations between the two entity mentions of interest ``{\it Acetazolamide} and ``{\it cyclosporine}''. In order to capture such important contexts for pooling, the typical approach in the prior work for RE is to perform the max-pooling operation over the abstract vectors for every word in the sentence (i.e., the whole set $A$) \cite{zeng2014Relation,dossantos2015Classifying,le2018Largescale}. The rationale is to select the features of the abstract vectors in $A$ with the highest values in each dimension to reveal the most important context for RE. The max-pooled vector over the whole set $A$ is combined with the $F_{ENT-ONLY}$ vector in this method:
  \begin{align*}
    F_{SENT} &= \operatorname{max-pool}{(a_1,a_2,\ldots,a_n)} \\
    F_{ENT-SENT} &= [F_{ENT-ONLY}, F_{SENT}]
  \end{align*}

3. {\it ENT-DYM}: Similar to {\it ENT-SENT}, this method also seeks the important context information beyond the two entity mentions of interest. However, instead of taking the whole vector sequence $A$ for the pooling, {\it ENT-DYM} divides $A$ into three separate vector subsequences based on the start and end indexes of the first and second entity mentions (i.e., $s_1$ and $e_2$) respectively. The max-pooling operation is then applied over these three subsequences and the resulting vectors are combined to form an overall vector (i.e., dynamic pooling) \cite{zeng2015Distant}:
  \begin{align*}
    F_{LEFT} &= \operatorname{max-pool}{(a_1,a_2,\ldots,a_{s_1-1})} \\
    F_{MIDDLE} &= \operatorname{max-pool}{(a_{s_1},a_{s_1+1},\ldots,a_{e_2})} \\
    F_{RIGHT} &= \operatorname{max-pool}{(a_{e_2+1},a_{e_2+2},\ldots,a_n)} \\
    F_{ENT-DYM} &= [F_{LEFT},F_{MIDDLE},F_{RIGHT},\\
    & F_{ENT-ONLY}]
  \end{align*}
%Note that we set $F_{LEFT}$ and $F_{RIGHT}$ to zero vectors if there is no context word in their respective ranges.

4. {\it ENT-DEP0}: The previous pooling methods have only relied on the sequential structures of the sentence where the chosen subsets of $A$ for pooling always contain vectors for the consecutive words in the sentence. Unfortunately, such sequential pooling might introduce irrelevant words into the selected subsets of $A$, potentially causing noise in the pooling features and impeding the performance of the RE models. For instance, in the previous sentence example ``{\it \underline{Acetazolamide} can elevate \underline{cyclosporine} levels.}'', the {\it ENT-SENT} and {\it ENT-DYM} methods woulds also include the word ``{\it levels}'' in the pooling subsets that is not very important for the relation prediction in this case. Consequently, in {\it ENT-DEP0}, we explore the possibility to use the dependency parse tree of the input sentence $W$ to filter out the irrelevant words for the pooling operation. In particular, instead of considering every word in the input sentence, {\it ENT-DEP0} only pools over the abstract vectors in $A$ that correspond to the words along the shortest dependency path (SDP) between the two entity mentions $M_1$ and $M_2$ in the dependency tree for $W$ (called $SDP0(M_1,M_2)$). Note that the shortest dependency paths have been shown to be able to select the important context words for RE in many previous work \cite{zhou2005Exploring,Chan:10,xu2016Improved}. Similar to {\it ENT-SENT} and {\it ENT-DYM}, we also include $F_{ENT-ONLY}$ in this method:
  \begin{align*}
    F_{DEP0} &= \operatorname{max-pool}_{a_i \in SDP0(M_1,M_2)}(a_i) \\
    F_{ENT-DEP0} &= [F_{DEP0},F_{ENT-ONLY}]
  \end{align*}

5. {\it ENT-DEP1}: This method is similar to {\it ENT-DEP0}. However, instead of directly pooling over the words in the shortest dependency path $SDP0(M_1,M_2)$, {\it ENT-DEP1} extends this path to also include every word that is connected to some word in $SDP0(M_1,M_2)$ via an edge in the dependency tree for $W$ (i.e., one edge distance from $SDP0(M_1,M_2)$). We denote this extended word set by $SDP1(M_1,M_2)$ for which the corresponding abstract vectors in $A$ would be chosen for the max-pooling operation. The motivation for $SDP1(M_1,M_2)$ is that the representations of the words close to the shortest dependency path between $M_1$ and $M_2$ might also provide useful information to improve the performance for RE. In our experiments, we find that one edge is the optimal distance to enlarge the shortest dependency paths. Using larger distance for the pooling mechanism would hurt the performance of the deep learning models for RE:
  \begin{align*}
    F_{DEP1} &= \operatorname{max-pool}_{a_i \in SDP1(M_1,M_2)}(a_i) \\
    F_{ENT-DEP1} &= [F_{DEP1},F_{ENT-ONLY}]
  \end{align*}

Once the overall representation vector $F$ for the input sentence $W$ and the two entity mentions of interest has been produced, we feed it into a feed-forward neural network with a softmax layer in the end to obtain the probability distribution $P(y|W,M_1,M_2) = \operatorname{feed-forward}(F)$ over the possible relation types for our RE problem. This probability distribution would then be used for both making prediction (i.e., by taking the relation type with the highest probability) and training models (i.e., by optimizing the negative log-likelihood function).

%% file: experiments.tex
\section{Experiments}

\subsection{Datasets}
%We conduct experiments on two biomedical datasets:

In order to evaluate the performance of the models in this work, we employ the following biomedical datasets for RE in the experiments:

%\begin{itemize}
%\item 

DDI-2013 \cite{herrero-zazo2013DDI}: This dataset contains 730 documents from the Drugbank database, involving about 25,000 examples for the training and test sets (each example consists of a sentence and two entity mentions of interest for classification). There are 4 entity types (i.e., \emph{drug}, \emph{brand}, \emph{group} and \emph{brand\_n}) and 5 relation types (i.e., \emph{mechanism}, \emph{advise}, \emph{effect}, \emph{int}, and \emph{no\_relation}) in this dataset. The \emph{no\_relation} is to indicate any example that does not belong to any relation types of interest. This dataset is severely imbalanced, containing 85\% negative examples in the training dataset. In order to deal with such imbalanced data, we employ weighted sampling that equally distributes the selection probability for the positive and negative examples.

%Following the prior work on this dataset \cite{chowdhury2013FBKirst,zhou2018PositionAware}, we report the micro-averaged F1 scores of the models on this dataset.

%The data set contains 730 documents and about 25k entity pairs (including test set) drawn from Drugbank database. It contains 4 entity types and represents 5 relation types (including negative \emph{no\_relation} type). The data set is severely imbalance, containing 85\% negative examples. For dealing with imbalanced data, \citet{chowdhury2013FBKirst} applied a two-phrase classification, in which one classifier detects positive instance and the other then classifies them. \citet{zhou2018PositionAware} used a binary softmax together with a multi-class softmax. In our experiment, we still treat our RE task as a multi-class classification, however, in order to mitigate the issue, we used weighted sampling, which make the appearance probability of positive and negative examples is equally 50\%. We report micro-averaged F1 scores on this data set as it is conventional.
%\item 

BB3 \cite{deleger2016Overview}. This dataset contains 95 documents; each of them involves a title and abstract from a document from the PubMed database. There are 800 examples in this dataset divided into two separate sets (i.e., the training set and the validation set). BB3 also include a test set; however, the relation types for the examples in this test set are not provided. In order to obtain the performance of the models on the test set, the performers need to submit their system outputs to an official API that would evaluate the output and return the model performance. We train the models in this work on the training data and employ the official API to obtain their test set performance to be reported in the experiments for this dataset.

%for this dataset is not provided; however, an official API is available to enable performers to submit their systems and receive the performance on the test set. We train the models in this work with the training set and send them to this API to obtain the corresponding performance.

%The data set is not severely imbalance, containing 61.4\% negative examples, so we did not use weighted sampling. We report micro-averaged F1 scores on this data set as it is conventional.  

%The data set contains 95 document text--titles and abstractions-and about 800 entity pairs (including test set) drawn PubMed database. The data set is not severely imbalance, containing 61.4\% negative examples, so we did not use weighted sampling. We report micro-averaged F1 scores on this data set as it is conventional.  

%\end{itemize}

Following the prior work on these datasets \cite{chowdhury2013FBKirst,lever2016VERSE,zhou2018PositionAware,le2018Largescale}, we use the micro-averaged F1 scores as the performance measure in the experiments to ensure a compatible comparison.

%Note that negative examples in both datasets are automatically generated by pairing all the entity mentions appearing in the same sentences that have not been annotated as positives. 

\subsection{Parameters and Resources}
%For each model, we use evolution algorithm to grid-search the best configuration (from scratch) with fixed sets of hyper-parameters. We run evolution algorithm 8 rounds, 8 candidates each round, only the best 3 candidates survive and mutate for next 8 candidates. We also continue to train and test our models with its best configuration for 4 another random seeds. We train each model for 70 and 50 epochs on DDI2013 and BB3 respectively.

As the DDI-2013 dataset does not involve a development set, we tune the parameters for the models in this work based on the validation data of the BB3 dataset and use the selected parameters for both datasets in the experiments. The best parameters from this tuning process include the learning rate of 0.5 and momentum of 0.8 for the stochastic gradient descent (SGD) optimizer with nesterov's momentum to optimize the models. In order to regularize the models, we apply dropout between layers with the drop rate for word embeddings set to 0.7 and other drop rates set to 0.5. We also employ the weight dropout {\it DropConnect} in \cite{wan2013Regularization} to regularize the hidden-to-hidden transition matrix within each bidirectional LSTM in the models \cite{merity2017Regularizing}. For all the models that involve bidirectional LSTMs (i.e., {\it BiLSTM}, {\it BiLSTM-CNN}, and {\it BiLSTM-GCNN}), two layers of bidirectional LSTMs are utilized with 300 hidden units for each LSTM network. For the models with CNN components (i.e., {\it CNN} and {\it BiLSTM-CNN}), we use one CNN layer with multiple window sizes of 2, 3, 4, and 5 for the filters (200 filters for each window size). For the {\it BiLSTM-GCN} model, two GCNN layers are employed with 300 hidden units in each layer. Finally, for the final feed-forward neural network to compute the probability distribution (i.e., $\operatorname{feed-forward}$), we utilize two hidden layers for which 1000 hidden units are used for the first layer and the number of hidden units for the second layer is determined by the number of relation types in the datasets.

\subsection{Evaluating the Pooling Methods for RE}

This section evaluates the performance of different pooling methods when they are applied to the deep learning models for RE on the two datasets DDI-2013 and BB3. In particular, we integrate each of the pooling methods in Section \ref{sec:pool} (i.e., {\it ENT-ONLY}, {\it ENT-SENT}, {\it ENT-DYM}, {\it END-DEP0}, and {\it END-DEP1}) into each of the deep learning models in Section \ref{sec:model} (i.e., {\it CNN}, {\it BiLSTM}, {\it BiLSTM-CNN}, and {\it BiLSTM-GCN}), resulting 20 different model combinations to be investigated in this section. For each model combination, we train five versions of the model with different random seeds for parameter initialization over the training datasets. The performance of such versions over the test sets is averaged to serve as the overall model performance on the corresponding dataset. Tables \ref{table:ddi_results} and \ref{table:bb3_results} report the performance of the models on the DDI-2013 dataset and BB3 dataset respectively.

\begin{table}[ht]
%\addtolength{\abovecaptionskip}{-2.0mm}
%\addtolength{\belowcaptionskip}{-3mm}
\small
  \centering
  \begin{tabular}{llll}
    \hline \textbf{Model} & \textbf{P} & \textbf{R} & \textbf{F1}
    \\ \hline 
     \multicolumn{4}{l}{{\it CNN}} \\
     + {\it ENT-ONLY}            & 52.7 & 43.1 & 47.4 \\
     + {\it ENT-SENT}            & 75.8 & 60.7 & 67.3 \\
     + {\it ENT-DYM}             & 66.5 & 70.6 & 68.5 \\
     + {\it ENT-DEP0}            & 59.8 & 61.5 & 60.6 \\
     + {\it ENT-DEP1}            & 67.6 & 65.1 & 66.3 \\
    \multicolumn{4}{l}{{\it BiLSTM}} \\
    + {\it ENT-ONLY}             & 74.0 & 69.4 & 71.6 \\
    + {\it ENT-SENT}             & 74.8 & 71.7 & 73.1 \\
    + {\it ENT-DYM}              & 71.5 & 73.4 & 72.4 \\
    + {\it ENT-DEP0}             & 72.8 & 69.4 & 71.1 \\
    + {\it ENT-DEP1}             & 71.6 & 76.4 & {\bf 73.9} \\
    % - entities-attn             & 72.2 & 71.6 & 71.9 & 1.0 \\
    \multicolumn{4}{l}{{\it BiLSTM-CNN}} \\
    + {\it ENT-ONLY}             & 69.6 & 72.3 & 70.9 \\
    + {\it ENT-SENT}             & 69.4 & 74.9 & 72.0 \\
    + {\it ENT-DYM}              & 71.0 & 69.7 & 71.8 \\
    + {\it ENT-DEP0}             & 72.2 & 69.5 & 70.8 \\
    + {\it ENT-DEP1}             & 71.0 & 74.3 & 72.6 \\
    % - entities-attn             & 65.8 & 69.3 & 67.5 & 0.8 \\
     \multicolumn{4}{l}{{\it BiLSTM-GCNN}} \\
    + {\it ENT-ONLY}             & 69.3 & 71.4 & 70.4 \\
    + {\it ENT-SENT}             & 72.2 & 71.9 & 72.0 \\
    + {\it ENT-DYM}              & 69.7 & 73.9 & 71.7 \\
    + {\it ENT-DEP0}             & 70.1 & 71.1 & 70.6 \\
    + {\it ENT-DEP1}             & 72.7 & 72.9 & 72.8 \\
    % - entities-attn             & 69.6 & 74.6 & 72.0 & 0.8 \\
    % \hline
    %\hline
    %\multicolumn{5}{l}{Average} \\
    %- entities-only             & 71.0 & 71.0 & 71.0 & - \\
    %- entities-sent             & 72.2 & 72.8 & 72.4 & - \\
    %- entities-dyna             & 70.7 & 72.3 & 71.5 & - \\
    %- entities-sdp0             & 71.7 & 70.0 & 70.8 & - \\
    %- entities-sdp1             & 71.8 & 73.9 & 73.1 & - \\
    
    %- entities-only             & 71.0 & 71.0 & 65.0 & - \\
    %- entities-sent             & 72.2 & 72.8 & 71.1 & - \\
    %- entities-dyna             & 70.7 & 72.3 & 70.7 & - \\
    %- entities-sdp0             & 71.7 & 70.0 & 68.3 & - \\
    %- entities-sdp1             & 71.8 & 73.9 & 71.4 & - \\

    \hline
  \end{tabular}
  \caption{Results on DDI 2013}
  \label{table:ddi_results} 
\end{table}

\begin{table}[ht]
%\addtolength{\abovecaptionskip}{-2.0mm}
%\addtolength{\belowcaptionskip}{-3mm}
\small
  \centering
  \begin{tabular}{llll}
    \hline \textbf{Model} & \textbf{P} & \textbf{R} & \textbf{F1}
    \\ \hline
    \multicolumn{4}{l}{{\it CNN}} \\
    + {\it ENT-ONLY}             & 54.2 & 65.7 & 59.1  \\
    + {\it ENT-SENT}             & 55.0 & 62.5 & 59.1  \\
    + {\it ENT-DYM}              & 54.6 & 53.3 & 53.5  \\
    + {\it ENT-DEP0}             & 55.9 & 65.8 & 60.6  \\
    + {\it ENT-DEP1}             & 55.7 & 67.7 & 61.1  \\
    \multicolumn{4}{l}{{\it BiLSTM}} \\
    + {\it ENT-ONLY}             & 58.9 & 59.6 & 59.2  \\
    + {\it ENT-SENT}             & 60.7 & 59.2 & 59.9  \\
    + {\it ENT-DYM}              & 50.2 & 66.0 & 56.9  \\
    + {\it ENT-DEP0}             & 51.6 & 78.0 & 61.9  \\
    + {\it ENT-DEP1}             & 54.7 & 72.6 & 62.4  \\
     \multicolumn{4}{l}{{\it BiLSTM-CNN}} \\
    + {\it ENT-ONLY}             & 56.4 & 66.2 & 60.8  \\
    + {\it ENT-SENT}             & 53.6 & 69.2 & 60.5  \\
    + {\it ENT-DYM}              & 47.1 & 78.0 & 58.7  \\
    + {\it ENT-DEP0}             & 55.9 & 71.4 & {\bf 62.5}  \\
    + {\it ENT-DEP1}             & 54.1 & 74.7 & 62.4  \\
    \multicolumn{4}{l}{{\it BiLSTM-GCNN}} \\
    + {\it ENT-ONLY}             & 62.7 & 56.1 & 58.9  \\
    + {\it ENT-SENT}             & 58.4 & 58.7 & 58.5  \\
    + {\it ENT-DYM}              & 56.8 & 58.4 & 56.6  \\
    + {\it ENT-DEP0}             & 55.6 & 67.4 & 60.8  \\
    + {\it ENT-DEP1}             & 54.4 & 71.1 & 61.5  \\
    %\hline
    %\multicolumn{5}{l}{Average} \\
    %- entities-only             & 59.4 & 60.1 & 59.7 & - \\
    %- entities-sent             & 58.3 & 62.0 & 59.6 & - \\
    %- entities-dyna             & 51.3 & 70.3 & 57.4 & - \\
    %- entities-sdp0             & 54.6 & 71.8 & 61.7 & - \\
    %- entities-sdp1             & 56.3 & 70.3 & 62.1 & - \\
    \hline
  \end{tabular}
  \caption{Results on BioNLP BB3}
  \label{table:bb3_results} 
\end{table}

From the tables, we have the following observations about the effectiveness of the pooling methods for RE with deep learning:

1. Comparing {\it ENT-SENT}, {\it ENT-DYM} and {\it ENT-ONLY}, we see that the pooling methods over the whole sentence (i.e., {\it ENT-SENT} and {\it ENT-DYM}) are significantly better than {\it ENT-ONLY} that only focuses on the two entity mentions of interest in the DDI-2013 dataset. This is true across different deep learning models in this work. However, this comparison is reversed for the BB3 dataset where {\it ENT-ONLY} is in general better or comparable to {\it ENT-SENT} and {\it ENT-DYM} over different deep learning models. We attribute such phenomena to the fact that the BB3 dataset often contains many entity mentions and relations within a single sentence (i.e., overlapping contexts) while the sentences in DDI-2013 tend to involve only a single relation with few entity mentions. This make {\it ENT-SENT} and {\it ENT-DYM}) ineffective for BB3 as the pooling mechanisms over the whole sentence are likely to involve the contexts for the other entity mentions and relations in the sentences, causing the low quality of the resulting representations and the confusion of the model for the relation prediction. This problem is less severe in DDI-2013 as the context of the whole sentence (with a single relation) is more aligned with the important context for the relation prediction. We call the many entity mentions and relations in a single single sentence of BB3 as the multiple relation effect for convenient discussion in this paper.

2. Comparing {\it ENT-SENT} and {\it ENT-DYM}, their performance are comparable in DDI-2013 (except for {\it CNN} where {\it ENT-DYM} is better); however, in the BB3 dataset, {\it ENT-SENT} singificantly outperforms {\it ENT-DYM} over all the models. This suggests the amplification of the multiple relation effect in BB3 due to {\it ENT-DYM} where the separation of the sentence context for pooling encourages the emergence of context information for multiple relations in the final representation vector and increases the confusion of the models.

3. Comparing the syntax-based pooling methods and the non-syntax pooling methods, the pooling based on dependency paths (i.e., {\it ENT-DEP0}) is worse than the non-syntax pooling methods (i.e., {\it ENT-SENT} and {\it ENT-DYM}) and perform comparably with {\it ENT-ONLY} in the DDI-2013 dataset over all the models (except for the {\it CNN} model where {\it ENT-ONLY} is much worse). These evidences suggest that the dependency paths themselves are not able to capture effective contexts for the pooling operation beyond the entity mentions for biomedical RE in DDI-2013. However, when we switch to the BB3 dataset, it turns out that {\it ENT-DEP0} is significantly better than all the non-syntax pooling methods (i.e., {\it ENT-ONLY}, {\it ENT-SENT} and {\it ENT-DYM}) for all the comparing models. This can be explained by the multiple relation effect in BB3 for which the dependency paths help to identify the most related context words for the two given entity mentions and filter out the confusing context words for the other relations in the sentences. The models would thus become less confused with different contexts for multiple relations as those in {\it ENT-SENT} and {\it ENT-DYM} for better performance in this case.

4. Finally, among all the pooling methods, we find that {\it ENT-DEP1} significantly outperforms the other pooling methods across different models and datasets (except the {\it CNN} model on DDI-2013 and {\it BiLSTM} on BB3). In particular, the performance improvement is substantial over the non-syntax pooling methods in BB3 where {\it ENT-DEP1} is up to 2\% better than {\it ENT-SENT}, {\it ENT-DYM} and {\it ENT-ONLY} on the absolute F1 scores. This helps to demonstrate the benefits of {\it ENT-DEP1} for biomedical RE to both recognize the important context words for pooling in DDI-2013 and reduce the confusion effect of the multiple relations in single sentences for the models in BB3.

\subsection{Comparing the Deep Learning Models for RE}

Regarding the comparison among different deep learning models, the major observations from from Tables \ref{table:ddi_results} and \ref{table:bb3_results} include:

1. The performance of {\it CNN} is in general worse that the other models with the bidirectional LSTM components (i.e., {\it BiLSTM}, {\it BiLSTM-CNN} and {\it BiLSTM-GCN}) over different pooling methods and datasets. This illustrates the importance of bidirectional LSTMs to capture the effective feature representations for biomedical RE. 
%For the {\it CNN} model, we also note that the pooling method {\it ENT-ONLY} is not 

2. Comparing {\it BiLSTM} and {\it BiLSTM-CNN}, we find that {\it BiLSTM} is better in DDI-2013 while {\it BiLSTM-CNN} achieves better performance in BB3 (over different pooling methods). In other words, the CNN layer is only helpful for the {\it BiLSTM} model in the BB3 dataset. This can also be attributed to the multiple relation effect in BB3 where the CNN layer helps to further abstract the representations from {\it BiLSTM} to better reveal the underlying structures in such confusing and complicated contexts in the sentences of BB3 for RE.

3. Graph convolutions over the dependency trees are not effective for biomedical RE as incorporating it into the {\it BiLSTM} model hurts the performance significantly. In particular, {\it BiLSTM-GCNN} is significantly worse than {\it BiLSTM} no matter which pooling methods are applied and which datasets are used for evaluation. 

4. Interestingly, comparing the {\it BiLSTM} model with the {\it ENT-DEP1} pooling method (i.e., {\it BiLSTM + ENT-DEP1}) and the {\it BiLSTM-GCN} model with the non-syntax pooling methods (i.e., {\it ENT-ONLY}, {\it ENT-SENT} and {\it ENT-DYM}), we see that {\it BiLSTM + ENT-DEP1} is significantly better with large performance gaps over both datasets DDI-2013 and BB3. For example, {\it BiLSTM + ENT-DEP1} is 1.9\% better than {\it BiLSTM-GCNN + ENT-SENT} in the DDI-2013 dataset and 3.5\% better than {\it BiLSTM-GCNN + ENT-ONLY} in BB3 with respect to the absolute F1 scores. In fact, {\it BiLSTM + ENT-DEP1} also achieves the best performance among the compared models in this section for both datasets. The major difference between {\it BiLSTM + ENT-DEP1} and {\it BiLSTM-GCN} with the non-syntax pooling methods lies at the specific component of the models where the syntactic information (i.e., the dependency trees) is applied. In {\it BiLSTM-GCN} with the non-syntax pooling methods, the syntactic information is employed in the representation learning component while in {\it BiLSTM + ENT-DEP}, the application of the syntactic information is postponed all the way to the pooling component. Our experiments thus demonstrate that it is more effective to utilize the syntactic information in the pooling component than in the representation learning component of the deep learning models for biomedical RE. This is an interesting and unique observation given that the prior work for RE has only focused on using the syntactic information in the representation component and never explicitly investigated the effectiveness of the syntactic information for the pooling component of the deep learning models.

\subsection{Comparing to the State-of-the-art Models}

In order to further demonstrate the advantage of the syntactic information for the pooling component for biomedical RE, this section compares {\it BiLSTM + ENT-DEP1} (i.e., the best model with the {\it ENT-DEP1} pooling in this work) with the best reported models on the two datasets DDI-2013 and BB3. For a fair comparison between models, we select the previous single (non-ensemble) models for the comparison in this section. Tables \ref{table:ddi_sota} and \ref{table:bb3_sota} presents the model performance.

\begin{table}[ht]
\centering
  \resizebox{0.48\textwidth}{!}{
    % \centering
    \begin{tabular}{llll}
      \hline \textbf{Models} & \textbf{P} & \textbf{R} & \textbf{F1}
      \\ \hline 
      \cite{raihani2017Rich}                   &\multirow{1}{*}{73.6} & \multirow{1}{*}{70.1} & \multirow{1}{*}{71.8}\\
      \cite{zhang2018DrugDrug}              &\multirow{1}{*}{74.1} & \multirow{1}{*}{71.8} & \multirow{1}{*}{72.9}\\
      \cite{zhou2018PositionAware}                       &\multirow{1}{*}{75.8} & \multirow{1}{*}{70.3} & \multirow{1}{*}{73.0}\\
      \cite{bjorne2018Biomedical}                            &\multirow{1}{*}{75.3} & \multirow{1}{*}{66.3} & \multirow{1}{*}{70.5}\\
      \hline
      %\multicolumn{4}{l}{{\it ENT-DEP1} pooling - this work} \\
      %{\it CNN}                & 67.6 & 65.1 & 66.3 \\
      %{\it BiLSTM}             & 71.6 & 76.4 & {\bf 73.9} \\
      {\it BiLSTM + ENT-DEP1}             & 71.6 & 76.4 & {\bf 73.9} \\
      %{\it BiLSTM-CNN}         & 71.0 & 74.3 & 72.6 \\
      %{\it BiLSTM-GCNN}        & 72.7 & 72.9 & 72.8 \\
      \hline
    \end{tabular}
  }
  \caption{Comparison with the state-of-the-art systems on the DDI-2013 test set}
  \label{table:ddi_sota}     
\end{table}

\begin{table}[ht]
\centering
  \resizebox{0.48\textwidth}{!}{
  \begin{tabular}{llll}
    \hline \textbf{Models} & \textbf{P} & \textbf{R} & \textbf{F1}
    \\ \hline 
    \cite{lever2016VERSE}                    & 51.0 & 61.5 & 55.8 \\
    \cite{mehryary2016Deep}               & 62.3 & 44.8 & 52.1 \\
    \cite{li2016Biomedical}                     & 56.3 & 58.0 & 57.1 \\
    \cite{le2018Largescale} & 59.8 & 51.3 & 55.2 \\
    \hline
    %\multicolumn{4}{l}{{\it ENT-DEP1} pooling - this work} \\
    %{\it CNN}                      & 55.7 & 67.7 & 61.1 \\
    %{\it BiLSTM}                   & 54.7 & 72.6 & {\bf 62.4} \\
    {\it BiLSTM + ENT-DEP1}                   & 54.7 & 72.6 & {\bf 62.4} \\
    %{\it BiLSTM-CNN}               & 54.1 & 74.7 & 62.4 \\
    %{\it BiLSTM-GCNN}              & 54.4 & 71.1 & 61.5 \\ 
    %\hline
  \end{tabular}
  }
  \caption{Comparison with the state-of-the-art systems on the BB3 test set}
  \label{table:bb3_sota} 
\end{table}

The most important observation from the tables is that the {\it BiLSTM} model, once combined with the {\it ENT-DEP1} pooling method, significantly outperforms the previous models on DDI-2013 and BB3, establishing new state-of-the-art performance for these datasets. In particular, in the DDI-2013 dataset, {\it BiLSTM + ENT-DEP1} is 0.9\% better than the current state-of-the-art model in \cite{zhou2018PositionAware} while the performance improvement over the best reported model for BB3 in \cite{li2016Biomedical} is 5.3\% (over the absolute F1 scores). Such substantial improvement clearly demonstrates the advantages of the syntactic information and its delayed application in the pooling component of the deep learning models for biomedical RE.

%For fair comparison, we selected recent single (non-ensemble) models that achieved state-of-the-art performance in \ref{table:ddi_sota} and \ref{table:bb3_sota}.
%On DDI2013 data set, with entities-sdp1 pooling approach, our BiLSTMs$\rightarrow$GCNs and BiLSTMs$\rightarrow$CNNs performs slightly worse than the state-of-the-art model with 0.3 and 0.4 fewer points of F1 score respectively. However, our BiLSTMs models outperforms the state-of-the-art model with 0.9 more points.
%On BB3 data set, with entities-sdp1 pooling approach, all three models significantly outperforms the state-of-the-art model with at least 4.4 more points of F1 score. Our BiLSTMs model achieves new best performance at F1 score 62.4.

%% file: related_conclusion.tex
\section{Related Work}

Traditional work on RE has mostly used feature engineering with syntactical information for statistical or kernel based classifiers \cite{zelenko2002Kernel,zhou2005Exploring,bunescu2005shortest,Sun:11,Chan:10}. Recently, deep learning has been shown to advance many benchmark datasets for this RE problem due to its representation learning capacity. The typical architectures for such deep learning models involve CNN, LSTM, the attention mechanism and their variants \cite{zeng2014Relation,dossantos2015Classifying,zhou2016AttentionBased,wang2016Relation,nguyen2015combining,miwa2016EndtoEnd,zhang2017Positionaware,zhang2018Graph}. Deep learning has also been applied to biomedical RE in the last couple of years and started to demonstrate much potentials for this area \cite{mehryary2016Deep,bjorne2018Biomedical, nguyen2018Convolutional,verga2018Simultaneously}.

Pooling is a common and crucial component in most of the deep learning models for RE. \cite{nguyen2015Relation,dossantos2015Classifying} apply the pooling operation over the whole sentence for RE while \citet{zeng2015Distant} proposes the dynamic pooling mechanism in the CNN models. However, none of these prior work systematically examines different pooling mechanisms for deep learning in RE as we do in this work.

\section{Conclusion}
We conduct a comprehensive study on the effectiveness of different pooling mechanisms for the deep learning models in biomedical relation extraction. Our experiments suggest that the pooling mechanisms have a significant impact on the performance of the deep learning models and a careful evaluation should be done to decide the appropriate pooling mechanism for the biomedical RE problem. From the experiments, we also find that syntactic information (i.e., dependency parsing) provides the best pooling methods for the models and biomedical RE datasets we investigate in this work (i.e., {\it ENT-DEP1}). We achieve the state-of-the-art performance for biomedical RE over the two datasets DDI-2013 and BB3 with such syntax-based pooling methods.

%In this paper, we conduct comprehensive study about pooling mechanisms and show
%that they have significant impact on the performance of deep learning RE models.
%We find that syntactic information from dependency parsing provide the best
%pooling strategies for biomedical RE; {\it ENT-DEP0} pooling not only alleviates
%the need for models involving around dependency structures but also give better
%performance for biomedical RE. With {\it ENT-DEP0} pooling, our BiLSTM model
%achieve new state-of-the-art on both DDI-2013 and BB3-2013 datasets, 73.9 F1 and
%62.4 F1 respectively.

% We also interpreted how the information flow in each of basic recurrent,
% convolution, and graph convolution layer assuming that feed-forward and
% gradient-backward propagation is executed correctly and efficiently.
% Incorporating dependency trees, the graph convolution layer has desired features
% from both convolution layer (weight-sharing mechanism) and recurrent layer
% (information-spreading mechanism). However, it performs slightly
% better than convolutional and performs worse than recurrent layer in our
% experiments. There are two conjectured challenges when applying this layer:
% we need a better dependency parser in biomedical domain, and we need a better
% mechanism to train multi-layer graph convolution efficiently. We leave these
% challenges in future work.